\documentclass[a4paper]{article}

\usepackage{INTERSPEECH2018}
\usepackage{multirow}
\usepackage{multicol}
\usepackage{cite}
\title{Minimum Bayes Risk Training of RNN-Transducer for End-to-End Speech Recognition}
\name{Chao Weng, Chengzhu Yu,  Jia Cui, Chunlei Zhang, Dong Yu}
\address{Tencent AI Lab, Bellevue, USA}
\email{cweng@tencent.com}

\begin{document}

\maketitle
\begin{abstract}

In this work, we propose minimum Bayes risk (MBR) training of RNN-Transducer (RNN-T) for end-to-end speech recognition. Specifically, initialized with a RNN-T trained model, MBR training is conducted via minimizing the expected edit distance  between the reference label sequence and on-the-fly generated N-best hypothesis. We also introduce a heuristic to incorporate an external neural network language model (NNLM) in RNN-T beam search decoding and explore MBR training with the external NNLM. Experimental results demonstrate an MBR trained model outperforms a RNN-T trained model substantially and further improvements can be achieved if trained with an external NNLM. Our best MBR trained system achieves absolute character error rate (CER) reductions of 1.2\% and 0.5\% on read and spontaneous Mandarin speech respectively over a strong convolution and transformer based RNN-T baseline trained on $\sim$21,000 hours of speech. 

\end{abstract}

\section{Introduction}

An end-to-end speech recognition system consolidates all necessary speech recognition components into one  compact neural framework which enables a simpler building process and allows a joint optimization among components. There are mainly three categories of end-to-end speech recognition systems, namely, connectionist temporal classification (CTC) based \cite{google-a2w,direct-a2w,ibm-a2w,tencent-a2w,iterated-ctc,deepspeech1}, attention based \cite{first-attention-asr,las,stateoftheart-las} and RNN-T based \cite{rnnt,rnnt-explore,streaming-rnnt,rnnt-mandarin,ms-rnnt}. RNN-T is first proposed in \cite{rnnt} to avoid the conditional independence assumption in CTC and achieved a state-of-the-art performance on a phoneme recognition task in \cite{rnnt-timit}. A RNN-T consists of three components, a transcription network (encoder), a prediction network (decoder) and a joint network. The transcription network encodes a sequence of inputs into hidden states and the prediction network takes previous emitted non-filler symbols as inputs and outputs prediction vectors. The joint network receives the encoded hidden states and prediction vectors as inputs and outputs a distribution over a set of target symbols plus an additional filler symbol “blank”. Similar to RNN-T, recurrent neural aligner (RNA) \cite{rna} is another sequence-to-sequence model proposed to remove conditional independence assumption in CTC. The difference between RNA and RNN-T is that the decoder (the prediction network) of a RNA is conditioned on both encoder outputs and previous emitted symbols (including blank) while the decoder of a RNN-T is only conditioned on previous non-blank symbols. The fact that the prediction network of a RNN-T is not conditioned on the encoder output allows the pre-training of the decoder as a NNLM on text-only data \cite{rnnt-explore}. 

MBR training, one of the most popular sequential discriminative training criteria, has been successfully applied in a hybrid speech recognition system such as minimum phone error (MPE) \cite{dnn_smbr,danthesis}, minimum word error (MWE) \cite{dnn_smbr,danthesis} and state level minimum Bayes risk (sMBR) \cite{danthesis,asgd-smbr,dnn_smbr,lstm_smbr}. Sampling-based discriminative training technique was also proposed \cite{embr,rna} to fine-tune a RNA model to minimize expected word error rate (WER). Recently, MBR training has been applied in an attention based end-to-end speech recognition system in \cite{mbr} and further improved in \cite{mbr-self}. However, there are few, if any, previous works on MBR training of RNN-T for end-to-end speech recognition.  

In this work, MBR training of RNN-T is proposed for end-to-end speech recognition. Specifically, initialized with a RNN-T trained model, MBR training is conducted via minimizing the expected edit distance  between the reference label sequence and on-the-fly generated N-best hypothesis. We also introduce a heuristic to incorporate an external neural network language models (NNLM) during beam search decoding and explore MBR training together with an external NNLM. Experimental results demonstrate an MBR trained model outperforms a RNN-T trained model substantially and further improvements can be achieved if trained with an external NNLM. Our best MBR trained system achieves absolute CER reductions of 1.2\% and 0.5\% on read and spontaneous Mandarin speech respectively over a strong convolution and transformer based RNN-T baseline trained on $\sim$21,000 hours of speech.

The remainder of the paper is organized as follows. In Section \ref{ssec:arch}, we first describe the convolution and transformer based RNN-T architecture adopted in this work. The gradient used in MBR training is derived in Section \ref{ssec:mbr}. In Section \ref{ssec:mbrlm}, we introduce a heuristic to incorporate an external NNLM during beam-search decoding and explain how to conduct MBR training with the incorporated external NNLM. All the experiment details and results are presented in Section 3. We conclude our work in Section 4.

\section{Minimum Bayes Risk Training of RNN-T}
\label{sec:mbr}

\subsection{Convolution and Transformer Based RNN-T}
\label{ssec:arch}
\begin{figure}[t]
  \centering
  \includegraphics[width=8cm]{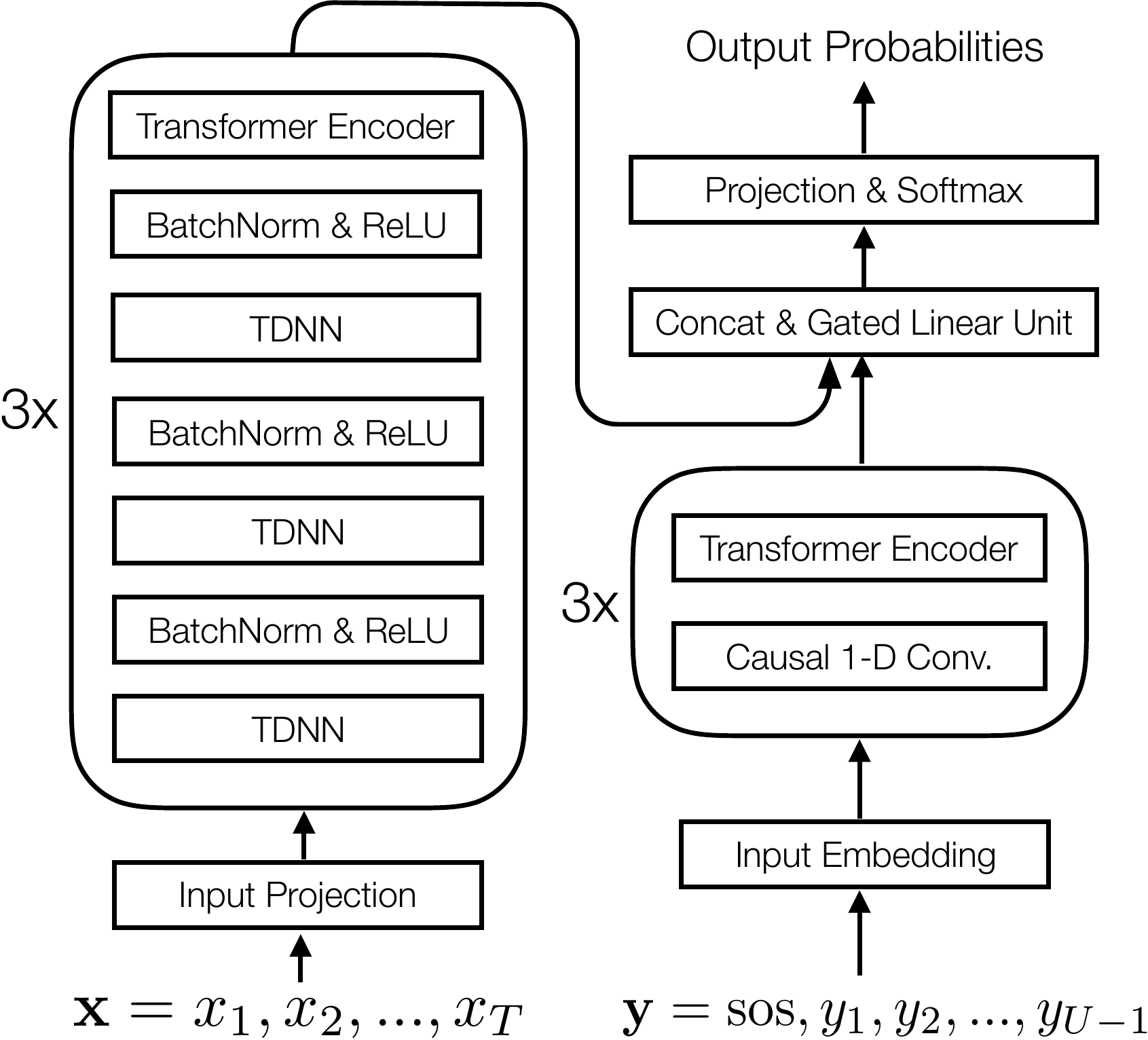}
  \caption{Convolution and transformer based RNN-T architecture adopted in this work.}
  \label{fig:rnnt_arch}
\end{figure}

Recently, transformer or self-attention based models have been proven very effective in both hybrid and end-to-end speech recognition systems \cite{transformer-las,transformer-hybrid,transformer-rnnt,fsmn-selfattention}. In this work, instead of an LSTM based architecture, a convolution and transformer based architecture is adopted for both encoder and decoder. The detailed architecture of RNN-T is depicted in Fig.\ref{fig:rnnt_arch}. The encoder is composed of three stacked blocks where each block consists of three time delay neural network (TDNN) layers \cite{tdnn} and one transformer encoder layer \cite{transformer}. The inputs to the encoder are speech features such as log-filterbanks or MFCCs extracted from the input speech signal. Given $T$ input speech frames $\mathbf{x} = x_1, x_2, ..., x_T $, the encoder transforms them into $\mathbf{h}^{\mathrm{enc}}$, a sequence of hidden states with the length $T$,
\begin{align}
\mathbf{h}^{\mathrm{enc}} &= h_1^{\mathrm{enc}}, h_2^{\mathrm{enc}}, ..., h_t^{\mathrm{enc}}, ...,  h_T^{\mathrm{enc}}  \nonumber \\
                          &= \mathrm{Encoder} (x_1, x_2, ..., x_t, ..., x_T). 
\label{eq:encoder}
\end{align}      
The decoder includes three basic blocks where each block contains one layer of 1-D causal convolution and one transformer encoder layer. It takes previous emitted labels $y_{1:u-1}$ as inputs and outputs the prediction vector $h_u^{\mathrm{dec}}$,
\begin{align}
h_u^{\mathrm{dec}} = \mathrm{Decoder}(y_{1:u-1}).
\label{eq:decoder}
\end{align}
Gated linear units \cite{glu} are adopted in the joint network as we found they improve the RNN-T performance significantly in our pilot experiments, 
\begin{align}
h_{t, u}^{\mathrm{joint}} = \mathrm{tanh}(W_f [h_t^{\mathrm{enc}}; h_u^{\mathrm{dec}}]) \oplus \sigma ( W_g [h_t^{\mathrm{enc}}; h_u^{\mathrm{dec}}]  ),
\label{eq:joint}
\end{align}
where $[a;b]$ denotes the concatenation of two vectors. $\sigma$ is the $\mathrm{sigmoid}$ function
and $\oplus$ is the element-wise product between tensors. Finally the projection and $\mathrm{softmax}$ layer produce the distribution of current target probabilities, 
\begin{align}
p(y_{t,u} | y_{1:u-1} , x_{1:t}) = \mathrm{softmax} (W_o h_{t,u}^{\mathrm{joint}}).
\label{eq:output}
\end{align}

\subsection{Minimum Bayes Risk Training of RNN-T}
\label{ssec:mbr}
Denote by $\mathbf{y}$ the output sequence from the RNN-T model (including blank), i.e., $\mathbf{y} = y_1, y_2, y_i, ..., y_{T+U}$ where $U$ is the number of non-blank symbols in the output sequence. Given $N$ pairs of the training speech utterance $\mathbf{x}$ and its corresponding reference label sequence $\mathbf{y}^r$ (note that $\mathbf{y}^r$ contains only non-blank symbols), the MBR loss function can be written as,
\begin{align}
\mathcal{L}_{\mathrm{MBR}}(\mathbf{x}_{1:N}, \mathbf{y}^r_{1:N}) = \sum_{n=1}^N \sum_{\mathbf{y}_n \in \mathcal{Y}_n} \frac{ P( \mathbf{y}_n | \mathbf{x}_n) R(\mathbf{y}_n, \mathbf{y}_n^r)}{\sum_{\mathbf{y}_n' \in \mathcal{Y}_n} P( \mathbf{y}_n' | \mathbf{x}_n)}, 
\label{eq:mbrobj}
\end{align}
where $\mathbf{y}_n$ and $\mathcal{Y}_n$ represent one of the hypothesis sequences and the whole hypothesis space correspond to $\mathbf{x}_n$. $R(\mathbf{y}_n, \mathbf{y}_n^r)$ is the risk function between a hypothesized and reference label sequence, e.g., edit-distance. $P( \mathbf{y}_n | \mathbf{x}_n)$ is the sequence probability given input $\mathbf{x}_n$ which is simply the product of $T+U$ output probabilities,
\begin{align}
P( \mathbf{y}_n | \mathbf{x}_n) &= \prod_{u=1}^{T+U} p(y_u | y_{1:u-1}, \mathbf{x}_n ). 
\end{align} 
Note that $p(y_u | y_{1:u-1}, \mathbf{x}_n)$ is exactly the output probabilities as in Eq. (\ref{eq:output}). Therefore, to perform MBR training, the gradients of MBR loss function w.r.t. $p(y_u | y_{1:u-1}, \mathbf{x}_n)$ need to be derived. For convenience, we use $p(y_u = y)$ as the shorthand for $p(y_u = y | y_{1:u-1}, \mathbf{x}_n)$, i.e., the probability of the model emitting a particular label $y$ at the $u$th step where $y$ could be one of the target or blank symbols. $f(\mathbf{y}_n)$ and $g(\mathbf{y}_n)$ are used as the shorthands for $P( \mathbf{y}_n | \mathbf{x}_n)$ and $R(\mathbf{y}_n, \mathbf{y}_n^r)$ in Eq. (\ref{eq:mbrobj}). Accordingly, we define a hypothesis set $S= \{\mathbf{y}_n | y_u = y \}$ which contains all the hypothesized sequences whose $u$th label equals to $y$. The whole hypothesis can be partitioned into two disjoint sets, i.e., $\mathbf{y}_n \in S$ and $\mathbf{y}_n \notin S$. The MBR loss function can be rewritten as, 
\begin{align}
\mathcal{L}_{\mathrm{MBR}} &= \sum_{n=1}^N \frac{ \sum_{\mathbf{y}_n \in S} f(\mathbf{y}_n) g(\mathbf{y}_n) + \sum_{\mathbf{y}_n \notin S} f(\mathbf{y}_n) g(\mathbf{y}_n) }{\sum_{\mathbf{y}_n' \in S} f(\mathbf{y}_n') +  \sum_{\mathbf{y}_n' \notin S} f(\mathbf{y}_n')},
\label{eq:mbrobjrewrite}
\end{align}
note that we have,
\begin{align}
 \frac{\partial \sum_{\mathbf{y}_n \notin S} f(\mathbf{y}_n)}{ \partial p(y_u = y)} &= 0, ~~~~~~ \frac{\partial \sum_{\mathbf{y}_n \notin S} f(\mathbf{y}_n)g(\mathbf{y}_n)}{ \partial p(y_u = y)} = 0
\end{align}
\begin{align}
 \frac{\partial \sum_{\mathbf{y}_n \in S} f(\mathbf{y}_n)}{ \partial p(y_u = y)} \cdot p (y_u=y) &= \sum_{\mathbf{y}_n \in S} f(\mathbf{y}_n),
\end{align}
\begin{align}
 \frac{\partial \sum_{\mathbf{y}_n \in S} f(\mathbf{y}_n)g(\mathbf{y}_n)}{ \partial p(y_u = y)} \cdot p (y_u=y) &= \sum_{\mathbf{y}_n \in S} f(\mathbf{y}_n)g(\mathbf{y}_n).
\end{align}
Therefore, the derivative of MBR loss function w.r.t. $\log p(y_u = y)$ can be written as,
\begin{align}
\frac{\partial \mathcal{L}_{\mathrm{MBR}}}{\partial \log p(y_u = y)} &=\sum_{n=1}^N \frac{\partial \mathcal{L}_{\mathrm{MBR}}}{\partial p(y_u = y)} \cdot \frac{\partial  p(y_u = y)}{\partial \log p (y_u = y) }  \nonumber \\
&= \sum_{n=1}^N \frac{\partial \mathcal{L}_{\mathrm{MBR}}}{\partial p(y_u = y)} \cdot  p(y_u = y) \nonumber \\
&= \sum_{n=1}^N \bigg ( \frac{\sum_{\mathbf{y}_n \in S} f(\mathbf{y}_n) g(\mathbf{y}_n)}{\sum_{\mathbf{y}_n'} f(\mathbf{y}_n')} - \nonumber \\
  & \frac{\sum_{\mathbf{y}_n} f(\mathbf{y}_n) g(\mathbf{y}_n) \sum_{\mathbf{y}_n \in S} f(\mathbf{y}_n)   }{[\sum_{\mathbf{y}_n'} f(\mathbf{y}_n')]^2} \bigg ) \nonumber \\
& = \sum_{n=1}^N \sum_{\mathbf{y}_n \in S}\gamma(\mathbf{y}_n) (g(\mathbf{y}_n) - \overline{R}_n),
\label{eq:mbrderive}
\end{align}
where $\gamma(\mathbf{y}_n)$ is the normalized sequence probability, i.e., $\gamma(\mathbf{y}_n) = \frac{ f(\mathbf{y}_n)}{\sum_{{\mathbf{y}}_n'} f(\mathbf{y}_n')} =  \frac{ P(\mathbf{y}_n | \mathbf{x}_n)} {\sum_{{\mathbf{y}}_n'} P(\mathbf{y}_n' |\mathbf{x}_n)}$. And $\overline{R}_n$ is the averaged risk among all the hypothesized sequences for the training utterance $n$,
\begin{align}
\overline{R}_n = \frac{\sum_{\mathbf{y}_n} f(\mathbf{y}_n) g(\mathbf{y}_n) }{\sum_{\mathbf{y}_n'} f(\mathbf{y}_n')} = \sum_{\mathbf{y}_n} \gamma(\mathbf{y}_n)  R(\mathbf{y}_n, \mathbf{y}_n^r).
\label{eq:mbraveragerisk}
\end{align}
As for the risk function, the edit-distance between the blank-removed hypothesis and the reference sequence is used in this work, i.e., 
\begin{align}
R(\mathbf{y}_n, \mathbf{y}_n^r) = \mathrm{Editdistance( \mathrm{RemoveBlank}(\mathbf{y}_n)  , \mathbf{y}_n^r)}
\end{align}
Since the forward computation of a RNN-T is conditioned on both acoustic inputs and previous emitted symbols, we use N-best from beam-search as the hypothesis space.

\subsection{Shallow Fusion and MBR Training with NNLMs}
\label{ssec:mbrlm}
Shallow fusion approach has been widely used in an attention based end-to-end speech recognition system \cite{shallowfusion} where the scores from an external LM are interpolated with model output scores at each step of beam search. However, unlike an attention based system, a RNN-T outputs additional blank symbols. To this end, we introduce a simple heuristic where the original blank scores are kept the same and the interpolation is only applied to those non-blank output scores. After such an interpolation, we normalize the scores for non-blank symbols to ensure that the total sum of the probabilities over all RNN-T outputs (including both blank and non-blank) equals to one. Specifically, 
\begin{equation}
\log p(y_u)= \left \{ \begin{array}{cc} \log \left [ p'(y_u) \frac{\sum_{y_u \neq \mathrm{blk}} p_{\mathrm{rnnt}} (y_u) }{\sum_{y_u \neq \mathrm{blk}} p'(y_u)} \right ] , & y_u \neq \mathrm{blk} \\
                                       \log p_{\mathrm{rnnt}} (y_u), & y_u = \mathrm{blk} \end{array} \right . ,
\label{eq:shallowfusion}
\end{equation}
where $ p'(y_u)$ is the interpolated probability of a non-blank symbol, 
\begin{equation}
\log p'(y_u) = (1-\lambda)\log p_{\mathrm{rnnt}} (y_u) + \lambda \log p_{\mathrm{nnlm}} (y_u).
\label{eq:shallowfusion1}
\end{equation}
Softmax smoothing \cite{Changhao-attention,mbr-self} is also apply to RNN-T outputs which we find significantly improves beam search. Specifically, Eq. (\ref{eq:output}) is modified to,  
\begin{align}
p_{\mathrm{rnnt}}(y_u) = \mathrm{softmax} (\beta W_o h_{t,u}^{\mathrm{joint}})
,~~~ \beta < 1.
\label{eq:softmaxsmoothing}
\end{align}         
To incorporate the external NNLM into MBR training, we simply conduct shallow fusion and beam search by employing Eq.\ref{eq:shallowfusion} and Eq.\ref{eq:softmaxsmoothing} to generate N-best hypothesis.   

\section{Experiments}
\label{sec:exp}
 All of our RNN-T systems are trained on $\sim$21,000 hours of transcribed Mandarin speech which includes $\sim$12,000 hours of read and $\sim$9,000 hours of spontaneous speech. We also augment acoustic training data three times using different speaking rates and volumes \cite{lfmmi}. Our models are evaluated on two test sets. One consists of about 1.5 hours of read speech (Read), the other one is composed of 2 hours of spontaneous speech (Spon). For input features, we use 40 dimensional high resolution MFCCs. The targets of our RNN-T system are a set of Mandarin characters and English letters plus blank symbol which leads to an output dimension of 6268. The text corpus used in external NNLM training contains about 80 billion of tokens. We use pytorch \cite{pytorch} and Kaldi \cite{kaldi} to implement all the models and experiments in this work.

\subsection{Detailed Configurations of RNN-T and NNLM}
See Fig.\ref{fig:rnnt_arch} for the architecture of RNN-T used in all our experiments. We also list detailed configurations of the encoder and decoder in Table \ref{tab:rnnt-enc-config} and Table \ref{tab:rnnt-dec-config}. The encoder contains 9 TDNN layers. The initial layers are trained on narrow contexts and the deeper layers receive the hidden activations from a wider temporal context which leads to an entire context of 21 left and 21 right frames at the last TDNN layer. Further, we set the stride of the last TDNN layer to 3 to do the subsampling which makes RNN-T training more memory efficient. After every three TDNN layers, we insert a transformer encoder layer where the input and output dimension is $d_{\mathrm{model}} = 1024$ and the inner-layer has also the same dimensionality $d_{\mathrm{ff}} = 1024$. The number of heads for self-attention is set to 16 for the first two transformer layers and 8 for the last transformer layer as the temporal length has been reduced by 3 after going through the last TDNN layer. The decoder includes 1 input embedding layer with an embedding dimension of 100 and 3 causal 1-D convolution layers with a kernel size of 5. After each causal convolution layer, a transformer layer is inserted to the decoder and a mask is employed to prevent the transformer layer from attending to those subsequent hidden activations. The number of total parameters in the RNN-T is about 65.09M.

As for the external NNLM, we adopt 2 layers of LSTM and each layer has 1024 hidden states. The embedding dimension is set to 200. The output targets correspond to the same set of non-blank symbols used in RNN-T with an additional start of sequence (SOS) symbol. The number of total parameters in the NNLM is about 22.13M. 
\begin{table}
\setlength{\tabcolsep}{3pt}
\centering
\caption{The detailed configuration of RNN-T encoder}
\begin{tabular}{c|cccc}
\hline
\hline
Enc. Layer &  Layer-type  & Context & $d_{\mathrm{hidden}}$ & stride  \\
\hline
1,2,3 & TDNN & [-1,0,1] & 1024 & 1  \\
5,6,7,9,10 & TDNN & [-3,0,3] & 1024 & 1  \\
11 & TDNN & [-3,0,3] & 1024 & 3 \\
\hline
Enc. Layer & Layer-type   & ($d_{\mathrm{model}}$, $d_{\mathrm{ff}}$) & \#heads & dropout  \\
     \hline
4, 8 &  Transformer & (1024, 1024) & 16  & 0.2 \\
12 & Transformer  & (1024, 1024) & 8  & 0.2 \\
\hline
\end{tabular}
\label{tab:rnnt-enc-config}
\end{table}

\begin{table}
\centering
\caption{The detailed configuration of RNN-T decoder}
\begin{tabular}{c|cccc}
\hline
\hline
Dec. Layer &   Layer-type & \#ch. (in, out) & kernel   \\
\hline
1 & Causal 1-D Conv. &  (100, 512)  &5 \\
3,5 &  Causal 1-D Conv. & (512, 512) &5  \\
\hline
Dec. Layer & Layer-type &  ($d_{\mathrm{model}}$, $d_{\mathrm{ff}}$) & \#heads    \\
\hline 
2,4,6 &  Transformer & (512, 2048) & 8  \\
\hline 
\end{tabular}
\label{tab:rnnt-dec-config}
\end{table}

\subsection{Distributed Training using BMUF with log-linear learning rate decay}
Both RNN-T and MBR training are conducted on 16 GPUs. Our distributed training strategy is based on block-wise model-update filtering (BMUF) with a Nesterov momentum scheme \cite{bmuf} but with different learning rate scheduling. With the original learning rate scheduling used in \cite{bmuf}, it is observed the training easily gets stuck in a local minimum. Instead, we set initial and final learning rates and specify a fixed number of training epochs before training (therefore there is no early stop and no validation set is used). After each sync period of BMUF, we adjust the learning rate according to,
\begin{align}
\mathrm{lr} := \mathrm{initial\_lr} * \exp \left (\frac{\mathrm{\#processed\_batches}}{\mathrm{\#total\_batches}} \log \frac{\mathrm{final\_lr}}{\mathrm{initial\_lr}} \right).
\end{align}
The sync period of BMUF is set to 5 batches through all our experiments. For both RNN-T and MBR training, we first discard any utterances that are longer than 12 seconds, sort the training utterances according to their length, group every N sequences, i.e., N is the batch size, and then shuffle the groups before training. For RNN-T training, the batch size is set to 8 while for MBR training we use 4 as the batch size with a beam size of 2 (N-best size) due to GPU memory limits. The initial and final learning rates are 1e-3 and 1e-4 respectively for RNN-T training and the training accomplishes after 5 sweeps of 3 times speed and volumes augmented data (this is comparable to 15 epoch training on non-augmented 21,000hrs of data). For MBR training, the initial and final learning rates are set to 1e-4 and 1e-6 and it is observed the training converges within one epoch of augmented training data. Both RNN-T and MBR training took about $\sim$12 days on 16 GPUs.

The same distributed training strategy is used to train a 2x1024 LSTM LM on an 80 billion token corpus. The initial and final learning rate are set to 1e-3 and 1e-4 respectively. The training converges within one sweep of the whole corpus. We use 128 as batch size and 64 as the number of backpropagation through time (BPTT) steps. The LM training took about $\sim$8 days on 7 GPUs. 
\subsection{RNN-T Loss Regularization for MBR Training}
In our pilot MBR training experiments, it is observed that without RNN-T loss regularization the N-Best generation (beam search) slows down significantly which reduced the training efficiency substantially. Therefore, the regularized MBR loss function is adopted,
\begin{align}
\mathcal{L}'_{\mathrm{MBR}} = \mathcal{L}_{\mathrm{MBR}} + \lambda \mathcal{L}_{\mathrm{rnnt}}, 
\label{mbrregularization}
\end{align}    
where $\lambda$ is the regularization factor which we set to $1.0$ throughout all our MBR experiments. Note that actual batch size w.r.t $\mathcal{L}_{\mathrm{MBR}}$ is enlarged by a factor of beam size used in N-best generation therefore $\lambda=1.0$ is a reasonable value and we did not tune $\lambda$ in our experiments. 

\subsection{Experimental Results}
\begin{table}
\centering
\begin{tabular}{c|c|c|c}
\hline 
\hline
\multirow{2}{*}{Methods} & \multirow{2}{*}{ Epoch} & \multicolumn{2}{c}{CERs(\%)}  \\
\cline{3-4}
  &  &  Read & Spon  \\
\hline
\hline
RNN-T & 1 & 8.8 & 18.1 \\ 
RNN-T &  2  & 7.0 & 16.5 \\
RNN-T &  3 & 6.6 & 16.0 \\
RNN-T &  4 & 6.5 & 15.7 \\
RNN-T &  5 & 6.5 & 15.6 \\
\hline
\end{tabular}
\caption{CERs of RNN-T trained models on Read and Spon evaluation sets}
\label{tab:rnnt-res-epochwise}
\end{table}
\begin{table}
\setlength{\tabcolsep}{3pt}
\centering
\begin{tabular}{c|c|c|c|c}
\hline 
\hline
\multirow{2}{*}{Methods} & \multirow{2}{*}{ Epoch} & softmax smoothing& \multicolumn{2}{c}{CERs(\%)}  \\
\cline{4-5}
  & & factor $\beta$ &  Read & Spon  \\
\hline
\hline
RNN-T &  5  & 1.0 & 6.5 & 15.6 \\
RNN-T &  5  & 0.8 & 6.2 & 15.4 \\
\hline
\end{tabular}
\caption{CERs of RNN-T trained models when decoding with different softmax smoothing factors}
\label{tab:rnnt-res-softmax}
\end{table}
\begin{table}
\setlength{\tabcolsep}{3pt}
\centering
\begin{tabular}{c|c|c|c|c}
\hline 
\hline
\multirow{2}{*}{Methods} & \multirow{2}{*}{+NNLM} &  \multirow{2}{*}{\#Parameters} & \multicolumn{2}{c}{CERs(\%)}  \\
\cline{4-5}
  &  & &  Read  & Spon  \\
\hline
\hline
RNNT & No & 65.09M  & 6.2 & 15.4  \\ 
RNNT & decode &65.09M + 22.13M & 5.7& 15.3  \\ 
+MBR& No & 65.09M & 5.6 & 15.1 \\
+MBR& decode & 65.09M + 22.13M & 5.2 & 15.2 \\
+MBR& train \& decode & 65.09M + 22.13M & 5.0 & 14.9 \\
\hline
\end{tabular}
\caption{CERs of RNN-T and MBR trained models when training and/or decoding with external NNLMs.}
\label{tab:rnnt-res-mbrlm}
\end{table}
 As aforementioned, the number of epochs for RNN-T training is set to 5. Each epoch of training data consists of 3 way speed and volumes perturbed data, i.e., $\sim$63,000 hours of speech. A beam size of 8 is used for decoding through all experiments. We listed CER results of RNN-T models after each epoch on Read and Spon evaluation sets in Table \ref{tab:rnnt-res-epochwise} which confirm the convergence of RNN-T training. Then we apply softmax smoothing during beam search decoding and list CER results in Table \ref{tab:rnnt-res-softmax}. The CERs on Read/Spon sets reduce from 6.5\%/15.6\% to 6.2\%/15.4\% when the softmax smoothing factor $\beta$ set to 0.8. In the following experiments, softmax smoothing factor is fixed to 0.8. With the externally trained NNLM, shallow fusion is conducted according to Eq.\ref{eq:shallowfusion} and Eq.\ref{eq:softmaxsmoothing} during beam search decoding. The first two rows of Table \ref{tab:rnnt-res-mbrlm} compare decoding results of RNN-T trained model with and without the external NNLM. The interpolation weight for the external NNLM, i.e., $\lambda$ in Eq.\ref{eq:shallowfusion1}, is set to 0.1. Shallow fusion with the externally trained NNLM further reduces CERs by 0.5\% and 0.1\% on Read and Spon sets respectively. We argue that the reason the performance on Read set improves more is that training corpus contains no spontaneously spoken transcription.
 
 Initialized with the best RNN-T trained model, two MBR training experiments are conducted, i.e., with and without the external NNLM. The 3rd row in Table \ref{tab:rnnt-res-mbrlm} shows the CER results of MBR trained model without using any external NNLM during training. It demonstrates that the MBR trained system without using an external NNLM during decoding already outperforms the baseline RNN-T system equipped with an NNLM during decoding. When decoding with an external NNLM, the MBR trained system improves on the Read set but unfortunately no performance gains are observed on the Spon set. Finally, further improvements can be achieved if the external NNLM is incorporated into both training and decoding of the MBR model as shown in the last row of Table \ref{tab:rnnt-res-mbrlm}. Overall the best MBR trained system when decoding with the external NNLM achieves an absolute CER reduction of 1.2\%/0.5\% on the Read and Spon sets respectively.

\section{Conclusions}
\label{sec:conclude}
In this work, we propose MBR training of RNN-T for end-to-end speech recognition. Specifically, initialized with a RNN-T trained model, MBR training is conducted via minimizing the expected edit distance  between the reference label sequence and on-the-fly generated N-best hypothesis. We also introduce a heuristic to incorporate an external NNLM in RNN-T beam search decoding and explore MBR training with the external NNLM. Experimental results demonstrate an MBR trained model outperforms a RNN-T trained model substantially and further improvements can be achieved if trained with an external NNLM. Our best MBR trained system achieves an CER reduction of 1.2\%/0.5\% on read and spontaneous Mandarin speech respectively over a strong convolution and transformer based RNN-T baseline trained on $\sim$21,000 hours of speech.

\bibliographystyle{IEEEtran}

\bibliography{mybib}

\begin{thebibliography}{10}
\providecommand{\url}[1]{#1}
\csname url@samestyle\endcsname
\providecommand{\newblock}{\relax}
\providecommand{\bibinfo}[2]{#2}
\providecommand{\BIBentrySTDinterwordspacing}{\spaceskip=0pt\relax}
\providecommand{\BIBentryALTinterwordstretchfactor}{4}
\providecommand{\BIBentryALTinterwordspacing}{\spaceskip=\fontdimen2\font plus
\BIBentryALTinterwordstretchfactor\fontdimen3\font minus
  \fontdimen4\font\relax}
\providecommand{\BIBforeignlanguage}[2]{{%
\expandafter\ifx\csname l@#1\endcsname\relax
\typeout{** WARNING: IEEEtran.bst: No hyphenation pattern has been}%
\typeout{** loaded for the language `#1'. Using the pattern for}%
\typeout{** the default language instead.}%
\else
\language=\csname l@#1\endcsname
\fi
#2}}
\providecommand{\BIBdecl}{\relax}
\BIBdecl

\bibitem{google-a2w}
H.~Soltau, H.~Liao, and H.~Sak, ``Neural speech recognizer: Acoustic-to-word
  {LSTM} model for large vocabulary speech recognition,'' \emph{CoRR}, vol.
  abs/1610.09975, 2016.

\bibitem{direct-a2w}
K.~Audhkhasi, B.~Ramabhadran, G.~Saon, M.~Picheny, and D.~Nahamoo, ``Direct
  acoustics-to-word models for english conversational speech recognition,'' in
  \emph{Interspeech 2017}, 2017, pp. 959--963.

\bibitem{ibm-a2w}
K.~Audhkhasi, B.~Kingsbury, B.~Ramabhadran, G.~Saon, and M.~Picheny, ``Building
  competitive direct acoustics-to-word models for english conversational speech
  recognition,'' \emph{CoRR}, vol. abs/1712.03133, 2017.

\bibitem{tencent-a2w}
\BIBentryALTinterwordspacing
C.~Yu, C.~Zhang, C.~Weng, J.~Cui, and D.~Yu, ``A multistage training framework
  for acoustic-to-word model,'' in \emph{Proc. Interspeech 2018}, 2018, pp.
  786--790. [Online]. Available:
  \url{http://dx.doi.org/10.21437/Interspeech.2018-1452}
\BIBentrySTDinterwordspacing

\bibitem{iterated-ctc}
G.~Zweig, C.~Yu, J.~Droppo, and A.~Stolcke, ``Advances in all-neural speech
  recognition,'' in \emph{{ICASSP} 2017}, 2017, pp. 4805--4809.

\bibitem{deepspeech1}
A.~Y. Hannun, C.~Case, J.~Casper, B.~Catanzaro, G.~Diamos, E.~Elsen,
  R.~Prenger, S.~Satheesh, S.~Sengupta, A.~Coates, and A.~Y. Ng, ``Deep speech:
  Scaling up end-to-end speech recognition,'' \emph{CoRR}, vol. abs/1412.5567,
  2014.

\bibitem{first-attention-asr}
J.~Chorowski, D.~Bahdanau, D.~Serdyuk, K.~Cho, and Y.~Bengio, ``Attention-based
  models for speech recognition,'' \emph{CoRR}, vol. abs/1506.07503, 2015.

\bibitem{las}
W.~Chan, N.~Jaitly, Q.~V. Le, and O.~Vinyals, ``Listen, attend and spell: A
  neural network for large vocabulary conversational speech recognition,'' in
  \emph{ICASSP}, 2016.

\bibitem{stateoftheart-las}
C.-C. Chiu, T.~N. Sainath, Y.~Wu, R.~Prabhavalkar, P.~Nguyen, Z.~Chen,
  A.~Kannan, R.~J. Weiss, K.~Rao, E.~Gonina, N.~Jaitly, B.~Li, J.~Chorowski,
  and M.~Bacchiani, ``State-of-the-art speech recognition with
  sequence-to-sequence models,'' \emph{CoRR}, vol. abs/1712.17169, 2017.

\bibitem{rnnt}
\BIBentryALTinterwordspacing
A.~Graves, ``Sequence transduction with recurrent neural networks,''
  \emph{CoRR}, vol. abs/1211.3711, 2012. [Online]. Available:
  \url{http://arxiv.org/abs/1211.3711}
\BIBentrySTDinterwordspacing

\bibitem{rnnt-explore}
\BIBentryALTinterwordspacing
K.~Rao, H.~Sak, and R.~Prabhavalkar, ``Exploring architectures, data and units
  for streaming end-to-end speech recognition with rnn-transducer,''
  \emph{CoRR}, vol. abs/1801.00841, 2018. [Online]. Available:
  \url{http://arxiv.org/abs/1801.00841}
\BIBentrySTDinterwordspacing

\bibitem{streaming-rnnt}
Y.~He, T.~N. Sainath, R.~Prabhavalkar, I.~McGraw, R.~Alvarez, D.~Zhao,
  D.~Rybach, A.~Kannan, Y.~Wu, R.~Pang, Q.~Liang, D.~Bhatia, Y.~Shangguan,
  B.~Li, G.~Pundak, K.~C. Sim, T.~Bagby, S.~Chang, K.~Rao, and A.~Gruenstein,
  ``Streaming end-to-end speech recognition for mobile devices,'' in
  \emph{{IEEE} International Conference on Acoustics, Speech and Signal
  Processing, {ICASSP} 2019, Brighton, United Kingdom, May 12-17, 2019}, 2019,
  pp. 6381--6385.

\bibitem{rnnt-mandarin}
\BIBentryALTinterwordspacing
S.~Wang, P.~Zhou, W.~Chen, J.~Jia, and L.~Xie, ``Exploring rnn-transducer for
  chinese speech recognition,'' \emph{CoRR}, vol. abs/1811.05097, 2018.
  [Online]. Available: \url{http://arxiv.org/abs/1811.05097}
\BIBentrySTDinterwordspacing

\bibitem{ms-rnnt}
J.~Li, R.~R. Zhao, H.~Hu, and Y.~Gong, ``Improving rnn transducer modeling for
  end-to-end speech recognition,'' \emph{ArXiv}, vol. abs/1909.12415, 2019.

\bibitem{rnnt-timit}
\BIBentryALTinterwordspacing
A.~Graves, A.~Mohamed, and G.~E. Hinton, ``Speech recognition with deep
  recurrent neural networks,'' \emph{CoRR}, vol. abs/1303.5778, 2013. [Online].
  Available: \url{http://arxiv.org/abs/1303.5778}
\BIBentrySTDinterwordspacing

\bibitem{rna}
H.~Sak, M.~Shannon, K.~Rao, and F.~Beaufays, ``Recurrent neural aligner: An
  encoder-decoder neural network model for sequence to sequence mapping,'' in
  \emph{INTERSPEECH}, 2017.

\bibitem{dnn_smbr}
K.~Vesel{\'{y}}, A.~Ghoshal, L.~Burget, and D.~Povey, ``Sequence-discriminative
  training of deep neural networks,'' in \emph{{INTERSPEECH} 2013,}, 2013, pp.
  2345--2349.

\bibitem{danthesis}
D.~Povey, ``Discriminative training for large vocabulary speech recognition,''
  Ph.D. dissertation, University of Cambridge, 2003.

\bibitem{asgd-smbr}
G.~Heigold, E.~McDermott, V.~Vanhoucke, A.~Senior, and M.~Bacchiani,
  ``Asynchronous stochastic optimization for sequence training of deep neural
  networks,'' in \emph{ICASSP 2014}, May 2014, pp. 5587--5591.

\bibitem{lstm_smbr}
A.~W. Senior, H.~Sak, F.~de~Chaumont~Quitry, T.~N. Sainath, and K.~Rao,
  ``Acoustic modelling with {CD-CTC-SMBR} {LSTM} {RNNS},'' in \emph{{ASRU}
  2015,}, 2015, pp. 604--609.

\bibitem{embr}
\BIBentryALTinterwordspacing
M.~Shannon, ``Optimizing expected word error rate via sampling for speech
  recognition,'' \emph{CoRR}, vol. abs/1706.02776, 2017. [Online]. Available:
  \url{http://arxiv.org/abs/1706.02776}
\BIBentrySTDinterwordspacing

\bibitem{mbr}
R.~Prabhavalkar, T.~N. Sainath, P.~N. Y.~Wu, Z.~Chen, C.~Chiu, and A.~Kannan,
  ``Minimum word error rate training for attention-based sequence-to-sequence
  models,'' in \emph{ICASSP}, 2018.

\bibitem{mbr-self}
\BIBentryALTinterwordspacing
C.~Weng, J.~Cui, G.~Wang, J.~Wang, C.~Yu, D.~Su, and D.~Yu, ``Improving
  attention based sequence-to-sequence models for end-to-end english
  conversational speech recognition,'' in \emph{Proc. Interspeech 2018}, 2018,
  pp. 761--765. [Online]. Available:
  \url{http://dx.doi.org/10.21437/Interspeech.2018-1030}
\BIBentrySTDinterwordspacing

\bibitem{transformer-las}
L.~Dong, S.~Xu, and B.~Xu, ``Speech-transformer: A no-recurrence
  sequence-to-sequence model for speech recognition,'' \emph{2018 IEEE
  International Conference on Acoustics, Speech and Signal Processing
  (ICASSP)}, pp. 5884--5888, 2018.

\bibitem{transformer-hybrid}
Y.~Wang, A.~Mohamed, D.~Le, C.~Liu, A.~Xiao, J.~Mahadeokar, H.~Huang,
  A.~Tjandra, X.~Zhang, F.~Zhang, C.~Fuegen, G.~Zweig, and M.~L. Seltzer,
  ``Transformer-based acoustic modeling for hybrid speech recognition,'' 2019.

\bibitem{transformer-rnnt}
C.-F. Yeh, J.~Mahadeokar, K.~Kalgaonkar, Y.~Wang, D.~Le, M.~Jain, K.~Schubert,
  C.~Fuegen, and M.~L. Seltzer, ``Transformer-transducer: End-to-end speech
  recognition with self-attention,'' 2019.

\bibitem{fsmn-selfattention}
Z.~You, D.~Su, J.~Chen, C.~Weng, and D.~Yu, ``Dfsmn-san with persistent memory
  model for automatic speech recognition,'' 2019.

\bibitem{tdnn}
V.~Peddinti, D.~Povey, and S.~Khudanpur, ``A time delay neural network
  architecture for efficient modeling of long temporal contexts,'' in
  \emph{INTERSPEECH}, 2015.

\bibitem{transformer}
\BIBentryALTinterwordspacing
A.~Vaswani, N.~Shazeer, N.~Parmar, J.~Uszkoreit, L.~Jones, A.~N. Gomez, L.~u.
  Kaiser, and I.~Polosukhin, ``Attention is all you need,'' in \emph{Advances
  in Neural Information Processing Systems 30}, I.~Guyon, U.~V. Luxburg,
  S.~Bengio, H.~Wallach, R.~Fergus, S.~Vishwanathan, and R.~Garnett, Eds.\hskip
  1em plus 0.5em minus 0.4em\relax Curran Associates, Inc., 2017, pp.
  5998--6008. [Online]. Available:
  \url{http://papers.nips.cc/paper/7181-attention-is-all-you-need.pdf}
\BIBentrySTDinterwordspacing

\bibitem{glu}
\BIBentryALTinterwordspacing
Y.~N. Dauphin, A.~Fan, M.~Auli, and D.~Grangier, ``Language modeling with gated
  convolutional networks,'' \emph{CoRR}, vol. abs/1612.08083, 2016. [Online].
  Available: \url{http://arxiv.org/abs/1612.08083}
\BIBentrySTDinterwordspacing

\bibitem{shallowfusion}
A.~Kannan, Y.~Wu, P.~Nguyen, T.~N. Sainath, Z.~Chen, and R.~Prabhavalkar, ``An
  analysis of incorporating an external language model into a
  sequence-to-sequence model,'' \emph{2018 IEEE International Conference on
  Acoustics, Speech and Signal Processing (ICASSP)}, pp. 1--5828, 2017.

\bibitem{Changhao-attention}
C.~Shan, J.~Zhang, Y.~Wang, and L.~Xie, ``Attention-based end-to-end speech
  recognition in mandarin,'' \emph{CoRR}, vol. abs/1707.07167, 2017.

\bibitem{lfmmi}
D.~Povey, V.~Peddinti, D.~Galvez, P.~Ghahremani, V.~Manohar, X.~Na, Y.~Wang,
  and S.~Khudanpur, ``Purely sequence-trained neural networks for {ASR} based
  on lattice-free {MMI},'' in \emph{Interspeech 2016,}, 2016, pp. 2751--2755.

\bibitem{pytorch}
A.~Paszke, S.~Gross, S.~Chintala, G.~Chanan, E.~Yang, Z.~DeVito, Z.~Lin,
  A.~Desmaison, L.~Antiga, and A.~Lerer, ``Automatic differentiation in
  pytorch,'' 2017.

\bibitem{kaldi}
D.~Povey, A.~Ghoshal, G.~Boulianne, N.~Goel, M.~Hannemann, Y.~Qian, P.~Schwarz,
  and G.~Stemmer, ``The kaldi speech recognition toolkit,'' in \emph{In IEEE
  2011 workshop}, 2011.

\bibitem{bmuf}
K.~{Chen} and Q.~{Huo}, ``Scalable training of deep learning machines by
  incremental block training with intra-block parallel optimization and
  blockwise model-update filtering,'' in \emph{2016 IEEE International
  Conference on Acoustics, Speech and Signal Processing (ICASSP)}, March 2016,
  pp. 5880--5884.

\end{thebibliography}


\end{document}